\documentclass{article}
\usepackage{spconf,amsmath,amsfonts,graphicx}
\usepackage{todonotes}
\usepackage[caption=false,font=footnotesize]{subfig}
\usepackage{varioref}               
\usepackage[bookmarks=true,bookmarksopen=true]{hyperref}               
\usepackage{cleveref}               

\hypersetup{
	colorlinks   = true,              
	urlcolor     = blue,              
	linkcolor    = blue,              
	citecolor    = blue                
}

\crefname{figure}{fig.}{figures}
\Crefname{figure}{Fig.}{Figures}
\newcommand{\pos}{\phantom{\neg}}

\title{THE WITS INTELLIGENT TEACHING SYSTEM: DETECTING STUDENT ENGAGEMENT DURING LECTURES USING CONVOLUTIONAL NEURAL NETWORKS}
\name{Richard Klein\sthanks{This work is based on the research supported in part by the National Research Foundation of South Africa for the grant, Unique Grant No. 94006.}, Turgay Celik$^{*\dagger}$}
\address{{$^*$School of Computer Science and Applied Mathematics, University of the Witwatersrand}\\ {Johannesburg, South Africa}\\{$^\dagger$School of Information Science and Technology, Southwest Jiaotong University} \\ {Sichuan Sheng 610031, China}}
\begin{document}
%
\maketitle
\begin{abstract}
To perform contingent teaching and be responsive to students' needs during class,  lecturers must be able to quickly assess the state of their audience.
While effective teachers are able to gauge easily the affective state of the students, as class sizes grow this becomes increasingly difficult and less precise.
The Wits Intelligent Teaching System (WITS) aims to assist lecturers with real-time feedback regarding student affect.
The focus is primarily on recognising engagement or lack thereof.
Student engagement is labelled based on behaviour and postures that are common to classroom settings.
These proxies are then used in an observational checklist to construct a dataset of engagement upon which a CNN based on AlexNet is successfully trained and which significantly outperforms a Support Vector Machine approach.
The deep learning approach provides satisfactory results on a challenging, real-world dataset with significant occlusion, lighting and resolution constraints.
\end{abstract}
\section{Introduction}
\label{sec:intro}

Modern technology has fundamentally changed the way that students can access information.
With online video lectures and intelligent tutoring systems that can provide a more customised learning experience, the utility of traditional \emph{chalk-and-talk} lectures must be reconsidered.
Chronic boredom and disengagement are problems identified in literature with many strategies proposed to help combat it \cite{astin1984student,larson1991boredom,shernoff2003student,machardy2012}.
While truly expert teachers are skilled at assessing the emotional state of the class and taking action to positively impact learning \cite{kapoor2007}, this is a difficult process and uncommon in many classrooms \cite{van2011patterns}.
Many of the strategies proposed to re-engage students revolve around interactivity.
But as class sizes increase, it becomes difficult both to identify disengaged students and to interact with them.

The Wits Intelligent Teaching System (WITS) aims to help lecturers address issues with disengagement by monitoring students during class and reporting engagement information back to the lecturer \cite{klein16wits,klein16im}.
The lecturer might then decide, for example, to change style, stand on the other side of the room, ask a specific group of students a question, or even give students a break.

\begin{figure}[t]
	\centering
		\includegraphics[clip=true,trim=0cm 1cm 0cm 1cm,width=0.95\linewidth]{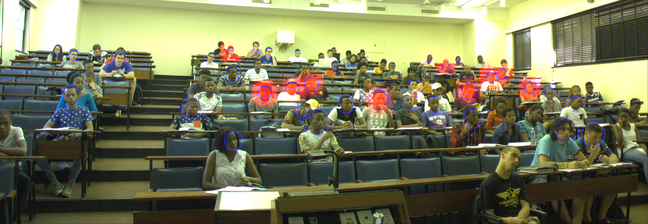}
		\caption{Interest Map}
		\label{fig:im}
\end{figure}

This data is presented to the lecturer through an Interest Map as seen in \Cref{fig:im} \cite{klein16im} .
The interest map highlights students that have disengaged which allows the lecturer to understand the spatial pattern of engagement in the class.
For example, a pervasive and uniform highlight over the class may indicate that the students generally are tired and a short break might be helpful for everyone.
On the other hand, if a specific part of the class is consistently highlighted by the system it may mean that the lecturer ignores that side of class or even that there is an acoustic, lighting or airflow problem.

Once interest maps can be generated reliably, various re-engagement strategies can be tested based on the different spatial distributions of interest.
Given this information, an \emph{Automatic Teaching Assistant (AutoTA)} could be created to monitor the interest map and suggest hints to the lecturer about what strategies are known to work in each circumstance.

In our previous works \cite{klein16wits,klein16im} we constructed a database of image sequences where students were labelled for common classroom actions as well as posture.
Using Histogram of Oriented Gradient features in conjunction a Support Vector Machine (HOG SVM), a classifier was built to try recognise these proxies over the dataset.
The HOG SVM could not generalise to new students very well. 
It was able to recognise writing with 65\% accuracy and cellphone use with about 60\% accuracy.
It was unable to beat chance when detecting other labels.

The proxies were used in an observational checklist to create labels of interest.
Interest maps were generated and shown to lecturers and professors at our institution.
Of 131 anonymous respondents, 81\% said they would like to receive live feedback about student interest during lectures, and 63\% found the interest map to be an intuitive representation of student interest (contrasted with 31\% neutral and 6\% unintuitive).
66\% of respondents said they would find an interest map useful to their teaching.
\begin{figure}[t]
	\centering
			\includegraphics[width=0.98\linewidth]{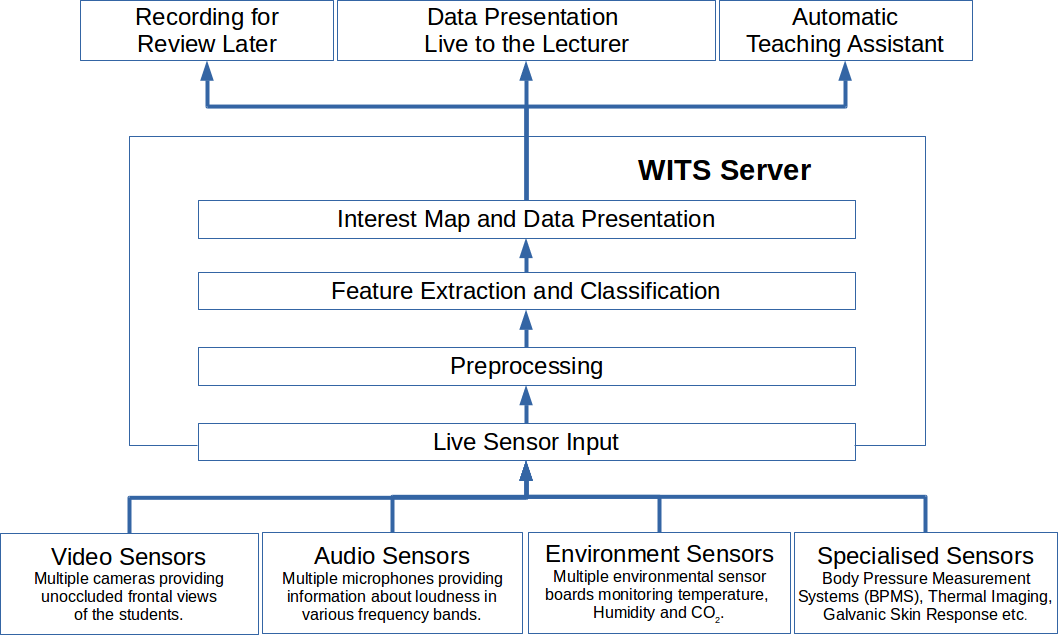}
			\caption{WITS System View}
			\label{fig:system}
\end{figure}

The HOG SVM approach suffers from significant accuracy problems when performing validation by holding out one subject at a time.
This may be a result of the HOG features not adequately capturing information in the video frames and with proper feature engineering an SVM may provide better results.
The benefit of Convolutional Neural Networks (CNNs), however, is that it learns a feature representation by itself and does not require an explicit feature engineering step on the part of the user.
The automatic feature learning coupled with fast CUDA enabled implementations like Caffe \cite{jia2014caffe} make CNNs an attractive choice.
This paper presents the results of a CNN based engagement classification strategy.

The rest of this paper is outlined as follows.
Section \ref{sec:wits} provides the architecture of WITS.
Section \ref{sec:data} explains the dataset and definition of engagement.
Section \ref{sec:cnn} briefly discuses the CNN architecture used in this paper.
Section \ref{sec:results} provides results and Section \ref{sec:conc} concludes the paper.

\section{WITS}\label{sec:wits}
The proposed architecture for WITS is shown in \Cref{fig:system}.
The smart lecture theatre will be equipped with various sensors capturing video, audio, and environmental data.
This is fed to the server that classifies students as engaged or not.
Once students have been classified, the information is passed to a presentation layer that builds the interest map according to some colour and transparency settings.
The interest map can then be saved for later review, displayed to the lecturer in real time, or passed to an AutoTA to suggest re-engagement strategies.
Currently the system only considers visual input but should be expanded in the future.

Our open source WITS Annotation Tool (\textsc{WITSat}) extracts manually labelled frames of individual students from lecture recordings.
Our previous work then extracted Histogram of Oriented Gradient features and performed dimensionality reduction using Principal Component Analysis.
A Support Vector Machine (SVM) was then trained on the labelled data to try recognise a number of classroom actions.
Validation was performed by holding out each student/subject one at a time: training on the remaining subjects and testing on the held out one.
It was partially able to recognise when students were writing and using their cellphone, but failed to generalise across students when detecting other actions in the dataset.
This paper attempts to recognise engagement by training on interest labels directly rather than attempting to first recognise the proxies.

\section{Data}\label{sec:data}
The database consists of image sequences from 3 lectures.
Lecture 1 and Lecture 2 took place on the same day with the same group of students.
Lecture 3 is the same group of students, but on a different day. 
Not all students are present in each lecture.

The data was labelled across common classroom actions and postures.
These actions include: writing, cellphone use, laptop use, talking, raised hand, yawning, and putting one's head on the desk.
Upper body posture labels include leaning left, right, backwards, forwards, and sitting up straight.
Head pose labels include looking far left or right, moderately left or right, below the desk, on the desk, upwards, or forwards.
Only one subject ever used a laptop in class and was therefore excluded from the experiments in this paper.

In the first attempt raters were asked to label engagement directly but due to poor inter-rater reliability these labels were unusable.
Therefore, an approximation is made by using the other labels in an observational checklist.
The definition of interest given here is limited by the data available and future work should focus on extending this definition by adding more labels based on the behaviours that lecturers encounter in the classroom.

The approach used here is a type of cascade, where each rule potentially rejects a sample as not engaged.
If a sample is never explicitly accepted or rejected, then it is labelled as engaged.
Let the label `Interest' indicate that a student is interested or engaged, and `$\neg$Interest' indicate that the student is not interested or disengaged.
The following rules about actions are considered:%
\begin{alignat}{2}
\mbox{Writing}      & \Rightarrow \pos \mbox{Interest},~  \mbox{Raised Hand}  && \Rightarrow \pos \mbox{Interest}\nonumber\\
\mbox{Cellphone}      & \Rightarrow \neg \mbox{Interest},~  \mbox{Head on Desk}    && \Rightarrow \neg \mbox{Interest}\nonumber\\
\mbox{Yawning}      & \Rightarrow \neg \mbox{Interest},~  \mbox{Talking} && \Rightarrow \neg \mbox{Interest}\nonumber
\end{alignat}
A fundamental assumption is that a student that is writing is actually taking notes and not just drawing or playing games.
A student with a raised hand is either asking a question or responding to one and therefore engaging with the lecturer.
On the other hand, a student that is on a cellphone or that has put their head on the desk has disengaged and lost interest.
Yawning is considered to be a symptom of both fatigue and boredom.
A student talking to a neighbour may be discussing work, asking a question, or talking about an upcoming party.
We are unable to tell the difference between these cases, so the assumption is that they are just \emph{chatting} and therefore disengaging.

The following rules about posture are considered:%
\begin{alignat}{2}
\mbox{Leaning Left} & \Rightarrow \neg \mbox{Interest},~ \mbox{Leaning Right} && \Rightarrow \neg \mbox{Interest}\nonumber
\end{alignat}
A student that is leaning far to either side is always disengaged.
This behaviour co-occurs with talking to friends, sleeping and cellphone use.
Those leaning only moderately to the side still exhibit a closed body posture that is considered either combative (resistance and aggression) or fugitive (withdrawal, defensive, boredom).
In both cases this implies disinterest or disengagement on the part of the student.
Those students with an open posture are usually sitting upright, leaning forwards or backwards.
Whether this indicates interest or not, largely depends on a student's head pose and actions.


The head pose labels mostly correspond to a student's focus of attention, and is interpreted as such because the resolution of the images does not allow for eye tracking.
These labels lead to the following rules involving head pose:%
\begin{alignat}{2}
\mbox{Far Left}  & \Rightarrow \neg \mbox{Interest},~ \mbox{Far Right} && \Rightarrow \neg \mbox{Interest}\nonumber\\
\mbox{Up}         & \Rightarrow \neg \mbox{Interest},~ \mbox{Below Desk}& & \Rightarrow \neg \mbox{Interest}\nonumber
\end{alignat}
Students looking to the far left, far right, or up at the roof are focusing neither on the lecturer nor on their notes.
They are either looking at a friend or looking around the lecture venue.
Sometimes it is the case that they are focusing on another student that is asking a question.
Detecting this would require a global consideration of the students in the video and should be the subject of future work.
Those that are looking below the desk are almost always distracted by something that they are trying to hide from the lecturer -- usually a cellphone.

Altogether these rules can be considered in a manner similar to the cascade classifier.
Firstly, the Writing and Raised Hand labels are checked -- if either is true, the student is marked interested.
The remaining rules are all used to reject students and mark them as $\neg$Interested.
If a student successfully passes through all `weak rules' without being rejected then the student is labelled as Interested.

\section{Convolutional Neural Network Architecture}\label{sec:cnn}
A Convolutional Neural Network (CNN) is a method whereby a neural network classifier is built upon a number of other convolution, pooling, drop-out, and normalisation layers.
The convolution layers learn the weights of a number of linear kernels to produce feature maps that are passed as input into the next layer.
Pooling layers summarise neighbouring groups of input neurons.
For example, max-pooling takes the maximum value as the summary of some neighbourhood.
By using overlapping neighbourhoods, over-fitting in the CNN is often reduced.
Normalisation layers can perform global, local, or lateral normalisation while drop-out layers help avoid over-fitting by stochastically deactivating neurons during the training process.
These layers are all used in conjunction with the fully connected layers typical of neural networks and training is performed using the usual back propagation algorithm.

AlexNet \cite{krizhevsky2012imagenet} is a CNN architecture that performed exceptionally well on the ImageNet database.
This work uses AlexNet to perform binary classification on the interest labels discussed in the previous section.
The final AlexNet softmax layer of 1,000 neurons is replaced with a 2 neuron softmax layer.

Image sequences are presented to the network as multiple channels of a single image.
This makes an image of $3k$ channels, where $k$ is the number of frames in the sequence.
By concatenating channels, pixels that are in the same location are `above' each other in the image cube and a kernel is able to easily compare pixels that correspond spatially.

\section{Results}\label{sec:results}
\subsection{Cross validation on a random subset of frames}\label{sec:res1}
From the available data, a balanced training set of 60,000 images (30,000 from each category) and a balanced testing set of 20,000 images (10,000 from each category) was sampled from the main dataset.
Batches of 256 images were used in each training epoch.
After 10,000 and 20,000 training epochs the CNN achieved 88.9\% and 89.7\% accuracy respectively.
After 100,000 iterations the accuracy was still at 89.8\% and had plateaued.
A stratified validation set of 200,000 images (89,765 negative, 110,235 positive) was randomly sampled from the remaining unused images to represent the correct proportions of images in the original data.
Over the whole validation set, it achieved 89.6\% accuracy.
The confusion matrix is shown in \Cref{tab:confusion-cnn1}.

\begin{table}[t]
	\centering
	\caption{CNN Confusion Matrix for the Validation Set}\label{tab:confusion-cnn1}
	\vspace{5pt}
	\begin{tabular}{|l|r|r|}
		\hline
		\multicolumn{1}{|c|}{Actual}  & \multicolumn{2}{c|}{Predicted}\\ \cline{2-3}
		&$\neg$Interested 	& Interested\\\hline \hline
		$\neg$Interested	&82,171		&13,149		\\\hline
		$\pos$Interested	&7,594		&97,086		\\\hline
	\end{tabular}
\end{table}

For comparison, a HOG SVM was trained on the same dataset.
Performing 5-fold cross validation over the training set provided an accuracy of 72.6\%.
Training the SVM over the entire training set yielded 73.0\% accuracy when tested on the stratified validation set.

%

\subsection{Cross subject validation}\label{sec:res2}
Many subjects were present across all three lectures.
Therefore, hold-one-out validation across subjects was performed using frames from lecture 1 only so that subjects would not be inadvertently repeated.
There are 50 labelled subjects in lecture 1.
For a selected test subject, the CNN was trained on the remaining 49 subjects.
A balanced training set would be built from 60,000 images selected at random from the 49 subjects so that there were 30,000 images in each category.
A balanced testing set was then built using an equal number of positive and negative images based on the number that were available for the current test subject.
A validation set was built using \emph{all} frames from the test subject.

50 CNNs (one for each subject holdout) were trained in this way to test how well the system generalised to unseen subjects.
On the balanced test sets, the minimum accuracy was 50.0\% (equivalent to chance), while the maximum was 91.5\%.
The mean and median accuracies were 65.6\% and 63.6\% respectively.

Classification on the unbalanced validation sets resulted in an accuracy of 59\% over all frames with the confusion matrix shown in \Cref{tab:confusion-cnn2}.
\begin{table}[t]
	\centering
	\caption{CNN Confusion Matrix for subject cross validation on all lecture 1 frames}\label{tab:confusion-cnn2}
	\vspace{5pt}
	\begin{tabular}{|l|r|r|}
		\hline
		\multicolumn{1}{|c|}{Actual}  & \multicolumn{2}{c|}{Predicted}\\ \cline{2-3}
		&$\neg$Interested 	& Interested\\\hline \hline
		$\neg$Interested	&100,535	&83,135		\\ \hline
		$\pos$Interested	&88,095		&141,141		\\\hline
	\end{tabular}
\end{table}


The HOG SVM trained and tested on the same data gives a minimum and maximum accuracy of  22.5\% and 77.8\%.
The mean and median accuracies are 49.4\% and 51.3\% respectively.
Accuracy over the entire validation set yields an accuracy of 53.5\%.
This is about 6\% lower than the CNN approach.
In total the validation included 441,756 images in total.

\subsection{Sequence length}
The previous two experiments used single images of students as the input to the classifier.
The experiment in \Cref{sec:res1} is repeated for sequences of 2 and 4 frames.
The results shown in \Cref{tab:cnn-accuracy} which show no significant improvement over the single frame case indicating that the CNN is able to extract the relevant information already.

\begin{table}[t]
	\centering
	\caption{CNN Test Set Accuracy by Sequence Length}\label{tab:cnn-accuracy}
	\vspace{5pt}
	\renewcommand{\tabcolsep}{3pt}
	\begin{tabular}{|c|c|c|}
		\hline
		Number of Frames	& Testing Accuracy  & Validation Accuracy\\ \hline \hline
		 1 & 89.8 & 89.6\\ \hline
		 2 & 89.7 & 89.4\\ \hline
		 4 & 90.0 & 89.7\\ \hline
	\end{tabular}
\end{table}

\subsection{Image Size}
The previous experiments were all performed on $64\times 64$ images.
Using single frame image sequences a number of larger image sizes were considered for the training, testing, and validation sets from \Cref{sec:res1}.
The results are shown in \Cref{tab:cnn-accuracy-2}.
The increased image size causes a decrease in accuracy. 
The cause is two-fold.
On the one hand the kernel size remains the same, so it can now only find smaller features relative to the whole image. 
On the other hand, as the size of the image increases the size of the feature maps does too.
This means the number of parameters in the model increases. 
Even with larger training sets of up to 240,000 images the accuracy using larger images remains similar to those above.

\begin{table}[t]
	\centering
	\caption{CNN Test Set Accuracy by Image Size}\label{tab:cnn-accuracy-2}
	\vspace{5pt}
	\begin{tabular}{|c|c|c|}
		\hline
		Image Size & Testing Accuracy & Validation Accuracy\\ \hline \hline
		$64\times 64$ & 89.8 & 89.6\\ \hline
		$96\times 96$ & 88.0 & 87.9\\ \hline 
		$128\times 128$ & 85.0  & 85.4    \\ \hline
	\end{tabular}
\end{table}

\section{Conclusion}\label{sec:conc}
This paper presents a Convolutional Neural Network (CNN) approach to recognising interest in students in large lecture venues in-the-wild.
In previous work students in lecturers were labelled according to engagement but inter-rater reliability was poor making the data unreliable.
The students were then labelled according to a number of proxy classroom actions and postures which were used in an observational checklist to identify engagement.
Results show that a CNN significantly out performs a Support Vector Machine trained on Histogram of Oriented Gradient (HOG SVM) features both in terms of actual accuracy on the data set (\Cref{sec:res1}) and in terms of generalisation capabilities (\Cref{sec:res2}).

By concatenating images by channel, temporal information was included but this extra information did not improve the accuracy of the CNN significantly.
Similarly, larger input images did not improve the accuracy and ultimately lead to a loss of generalisation capabilities.

Using CUDA enabled GPU hardware, the CNN in Caffe \cite{jia2014caffe} vastly outperforms a parallel CPU implementation of HOG feature extraction and SVM classification with libSVM \cite{chang2011libsvm}.
With enough computational power and correctly implemented parallelism, the CNN based classifiers can perform this interest classification in real-time and interest information can be reported back to the presenter to help improve lectures through the use of an Interest Map or, in the future, an Automatic Teaching Assistant.


\bibliographystyle{IEEEbib}
\bibliography{references}

\end{document}